\documentclass[conference]{IEEEtran}
\IEEEoverridecommandlockouts
\usepackage{cite}
\usepackage{amsmath,amssymb,amsfonts}
\usepackage{algorithmic}
\usepackage{graphicx}
\usepackage{textcomp}
\usepackage{xcolor}
\usepackage{array}
\usepackage{tabularx}
\usepackage{booktabs}
\usepackage{url}
\usepackage{multirow}
\def\BibTeX{{\rm B\kern-.05em{\sc i\kern-.025em b}\kern-.08em
    T\kern-.1667em\lower.7ex\hbox{E}\kern-.125emX}}
\begin{document}

\title{PowerYOLO: Mixed Precision Model for Hardware Efficient Object Detection with Event Data
\thanks{The work presented in this paper was supported by: the program ''Excellence initiative –- research university'' for the AGH University of Krakow. We gratefully acknowledge Polish high-performance computing infrastructure PLGrid (HPC Centers: ACK Cyfronet AGH) for providing computer facilities and support within computational grant no. PLG/2023/01619}
%\thanks{Removed for blind review}
}

\author{\IEEEauthorblockN{Dominika Przewlocka-Rus\IEEEauthorrefmark{1},
Tomasz Kryjak\IEEEauthorrefmark{2} \textit{Senior Member IEEE} and
Marek Gorgon\IEEEauthorrefmark{3}  \textit{Senior Member IEEE}}
\IEEEauthorblockA{Embedded Vision Systems Group,
AGH University of Krakow\\
Krakow, Poland\\
Email: \IEEEauthorrefmark{1}dprze@agh.edu.pl,
\IEEEauthorrefmark{2}tomasz.kryjak@agh.edu.pl,
\IEEEauthorrefmark{3}mago@agh.edu.pl}}
%\author{Removed for blind review}

% \author{\IEEEauthorblockN{Dominika Przewlocka-Rus}
% \IEEEauthorblockA{\textit{Deptartament of Automatic Control and Robotics} \\
% \textit{AGH University of Krakow}\\
% Krakow, Poland \\
% \texttt{dprze@agh.edu.pl}}
% \and
% \IEEEauthorblockN{Tomasz Kryjak, Senior Member IEEE}
% \IEEEauthorblockA{\textit{Department of Automatic Control and Robotics} \\
% \textit{AGH University of Krakow}\\
% Krakow, Poland \\
% \texttt{tomasz.kryjak@agh.edu.pl}}
% \and
% \IEEEauthorblockN{Marek Gorgon}
% \IEEEauthorblockA{\textit{Department of Automatic Control and Robotics} \\
% \textit{AGH University of Krakow}\\
% Krakow, Poland \\
% \texttt{mago@agh.edu.pl}}
% }

\maketitle

\begin{abstract}
The performance of object detection systems in automotive solutions must be as high as possible, with minimal response time and, due to the often battery-powered operation, low energy consumption.
When designing such solutions, we therefore face challenges typical for embedded vision systems: the problem of fitting algorithms of high memory and computational complexity into small low-power devices.
In this paper we propose PowerYOLO -- a mixed precision  solution, which targets three essential elements of such application.
First, we propose a system based on a Dynamic Vision Sensor (DVS), a novel sensor, that offers low power requirements and operates well in conditions with variable illumination.
It is these features that may make event cameras a preferential choice over frame cameras in some applications.
Second, to ensure high accuracy and low memory and computational complexity, we propose to use 4-bit width Powers-of-Two (PoT) quantisation for convolution weights of the YOLO detector, with all other parameters quantised linearly.
Finally, we embrace from PoT scheme and replace multiplication with bit-shifting to increase the efficiency of hardware acceleration of such solution, with a special convolution-batch normalisation fusion scheme.
The use of specific sensor with PoT quantisation and special batch normalisation fusion leads to a unique system with almost 8x reduction in memory complexity and vast computational simplifications, with relation to a standard approach.
This efficient system achieves high accuracy of mAP $0.301$ on the GEN1 DVS dataset, marking the new state-of-the-art for such compressed model.
\end{abstract}

\begin{IEEEkeywords}
embedded vision systems, dynamic vision sensor, hardware-aware algorithm desing, logarithmic quantization, mixed precision, pedestrian detection, power-of-two quantization, vehicle detection,
\end{IEEEkeywords}

\section{Introduction}
\label{sec:introduction}

Vision systems based on machine learning algorithms are already becoming a standard in many areas of our lives: in applications and devices that provide entertainment (\textit{intelligent} photo and video processing, augmented/virtual reality), in surveillance and security systems and autonomous vehicles, or advanced driver assistance systems.

Naturally, in most applications we consider devices with limited energy budget -- mobile and battery-powered -- which means that the use of power-hungry algorithms based on neural networks requires careful design of such solutions, generally with some reduction in memory-computational complexity relative to standard approaches.

%Systemy wizyjne oparte o algorytmy uczenia maszynowego stanowią już pewien standard w przestrzeni automatyzacji wielu obszarów naszego życia: w aplikacjach i urządzeniach dostarczających rozrywki (\textit{inteligentna} obróbka zdjęć i wideo, rozszerzona/wirtualna rzeczywistość), w systemach nadzoru i bezpieczeństwa oraz autonomicznych pojazdach, lub systemach wspomagających kierowców.
%Naturalnie, w większości zastosowań mówimy o urządzeniach z ograniczonym budżetem energetycznym - mobilnych i zasilanych akumulatorowo -, co oznacza że użycie power-hungry algorytmów opartych o sieci neuronowe wymaga ostrożnego projektowania takich rozwiązań, z reguły z pewną redukcją złożoności pamięciowo-obliczeniowej względem standardowego podejścia.

Such a reduction generally means changing the precision of computations from 32-bit floating-point numbers to 16-bit floating-point, 8-bit fixed-point or even, in special cases, 6-bit and 4-bit fixed-point numbers, but above all to 8-bit or even lower-width integers, including binary and ternary.
With the right hardware platform, such simplifications can lead to increased performance (GPU, CPU), often also allowing a reduction in energy requirements with the right design of custom processor (FPGA, ASIC).
In addition, they allow to address to problem of memory bandwidth gap and reduce the size of the model. 
It is a standard practice to use quantisation to 8-bit integers, which only slightly affects the effectiveness of the neural network, especially if appropriate quantisation aware training (QAT) is used.

%Taka redukcja z reguły oznacza zmniejszenie precyzji obliczeń z liczb zmiennoprzecinkowych 32-bitowych do zmiennoprzecinkowych 16-bitowych, 8-bitowych lub nawet w specjalnych przypadkach 6-cio i 4-bitowych, liczb stałoprzecinkowych, ale przede wszystkim do liczb całkowitoliczbowych 8-bitowych lub jeszcze niższej szerokości, w tym binarnych i tenarnych.
%Przy odpowiednio dobranej platformie sprzętowej takie uproszczenia mogą prowadzić do zwiększenia wydajności (GPU, CPU), często pozwalając też na redukcję zapotrzebowania energetycznego przy odpowiednim zaprojektowaniu procesora przetwarzającego dane (FPGA, ASIC).
%Ponadto pozwalają zaadresować problem memory bandwidth gap, a także zmieejszyć rozmiar modelu.
%Standardowo przyjęło się użycie kwantyzacji do 8-bitowych liczb całkowitych, która w sposób jedynie nieznaczny wpływa na skuteczność sieci neuronowejm szczególnie jeśli zastosuje się odpowiednie doternowywanie modelu -- Post Quataztion Training.

The use of lower precision while retaining the accuracy of the full precision model may require specific quantisation schemes, such as the logarithmic quantisation benchmarked earlier for the classification task in \cite{b1}, \cite{b2}, \cite{b3}.
%Zastosowanie niższej precyzji zachowując skuteczność modelu pełnej precyzji może wymagać specyficznych schematów kwantyzacji, jak np. kwantyzacji logarytmicznej, benchmarkowanej wcześniej dla zadania klasyfikacji \cite{b1}, \cite{b2}, \cite{b3}.

Pedestrian and vehicle detection are tasks present in a wide range of applications: from driver assistance systems and autonomous vehicles, to security and surveillance systems.
All those applications have similar requirements -- high performance, real-time data processing, and rather low energy budget.
Therefore, the design of such a solution must be approached comprehensively: by choosing the right algorithm, hardware platform and even sensor.
A standard digital camera is a natural choice, while an event camera (dynamic vision sensor, neuromorphic camera) \cite{b4} is recently becoming an interesting alternative.

In their operation, event cameras mimic the characteristics of the human visual system, noticing only changes in brightness per pixel (and therefore events per pixel).
As a result, they record and transmit data with lower latency than standard cameras and also have a higher dynamic range, making them work well in a wide variety of lighting conditions, including unevenly and poorly illuminated scenes.
The latter can be particularly important in changing road conditions.

\begin{figure}[!t]
\centerline{\includegraphics[width=0.4\textwidth]{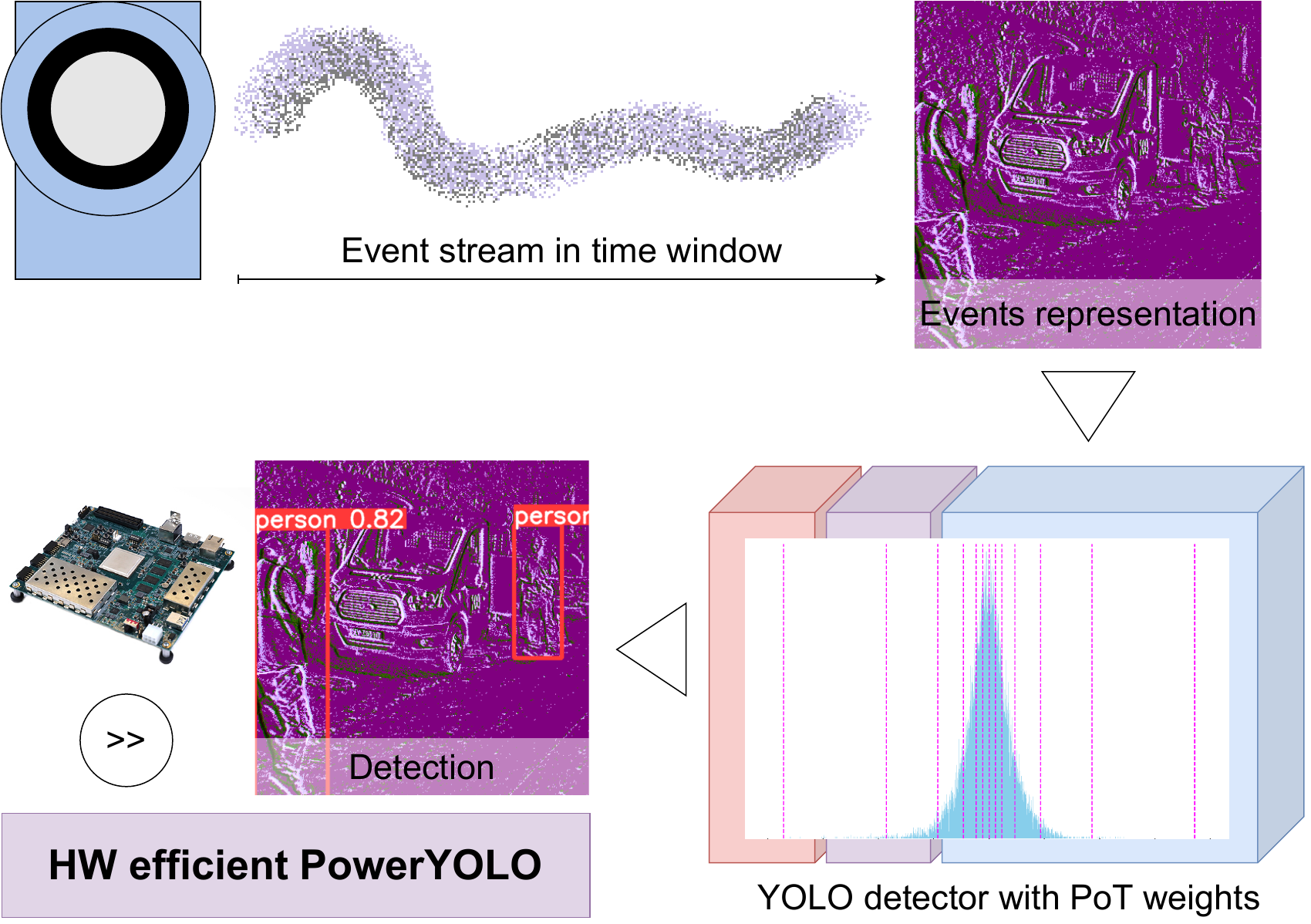}}
  \caption{Overview of the proposed system: event camera data is gathered over a predetermined time window (10 ms), and represented in the form of a pseudo-image (event frame). Pedestrians are detected using a mixed precision YOLO network, with convolution layer weights quantised logarithmically to PoT values, and the rest of parameters and activations quantised linearly.}
  \label{fig:imabs}
\end{figure}

%Detekcja pieszych i pojazdów jest jednym z tych systemów, które znajdują zastosowanie w wielu aplikacjach: od systemów wspomagania kierowców, przez autonomiczne pojazdy, aż do systemów bezpieczeństwa i monitoringu.
%Wszystkim stawiamy podobne wymagania - wysoką skuteczność działania, przetwarzanie danych w czasie rzeczywistym, oraz z niskim zapotrzebowaniem energetycznym.
%Dlatego też do projektowania takiego rozwiązania należy podejść kompleksowo: wybierając odpowiedni algorytm, platformę sprzętową, a nawet czujnik.
%Naturalnym wyborem jest standardowa kamera cyfrowa, natomiast interesującą alternatywę stanowi kamera zdarzeniowa (Dynamic Vision Sensor, neromorficzna).
%Kamery zdarzeniowe w swoim działaniu niejako naśladują cechy układu wzrokowego człowieka, rejestrując jedynie zmiany jasności per piksel (a więc zdarzenia per piksel).
%W konsekwencji rejestrują i transmitują dane z niższą niż standardowe kamery latencją, a także charakteryzują się wysoką rozpiętością tonalną, co sprawia, że dobrze działają w różnorodnych warunkach oświetleniowych, w tym scenach nierównomiernie i słabo oświetlonych.
%To ostatnie może być szczególnie istotne w zmiennych warunkach drogowych.

% In this paper we propose experiments that explore the use of logarithmic quantisation for tasks other than classification, in particular for a task customarily considered in the context of low-power devices, as in the case of autonomous vehicles -- pedestrian and vehicle detection.
In this paper we propose to use logarithmic quantisation for a task complex and commonly considered in the context of low-power devices -- pedestrian and vehicle detection.
In addition to designing an algorithm with low memory and computational complexity, it was also decided to use an event camera which characteristics are perfectly suited to the real-time and high accuracy requirements of embedded perception systems.
A schematic of the system is presented in Fig. \ref{fig:imabs}: the data received from the event camera is represented by a pseudo-image (i.e. event frame), which is then processed by the PowerYOLO mixed-precision network. 
Due to the PoT weights, a hardware implementation of the PowerYOLO network can introduce significant energy gains, resulting from the simplification of multiplication operations and therefore reducing the number of required electronic components needed to implement the corresponding accelerator.
Thus, the main contribution of this work can be summarised as follows:
\begin{enumerate}
    \item we propose a highly efficient INT8/LOG4 mixed precision PowerYOLO network for pedestrian and vehicle detection for GEN1 event dataset. Our model stands out for having the best ratio of efficiency to memory-computation complexity achieving mAP 0.301.
    \item we propose a method for convolution and batch normalisation layers fusion in a way that takes into account the special form of the neural network weights when quantised to powers of two.
    %\item Symulacja sprzętowa warstwy konwolucyjnej PoT w framework'u FINN i porównanie z liniowo kwantyzowaną warstwą. 
\end{enumerate}

The reminder of this paper is organised as follows:
Section \ref{sec:event_cameras} presents basic information about event cameras.
Section \ref{sec:related_work} discusses previous work related to the optimisation of neural network architectures and an overview of object detection methods based on event data stream.
Section \ref{sec:eyolo} presents the PowerYOLO network, which uses a mixed integer and power-of-two quantisation scheme.
Its evaluation is then provided is Section \ref{sec:results}.
The article concludes with a brief summary and suggestions for further research.

% -----------------------------------------------------------------------------------------
\section{Event cameras}
\label{sec:event_cameras}

Event cameras record changes in light intensity individually (asynchronously) for each pixel.
A single event is represented by a vector $e = \{t, x, y, p\}$, where $t$ is the occurrence time of the event (timestamp), $x$ and $y$ are the spatial coordinates of the recorded pixel, and $p$, called polarisation, takes the value $-1$ or $1$ (sometimes also $0$ and $1$) depending on the direction of brightness change.
Therefore, data acquisition only takes place when the brightness of the pixel changes, which enables high-resolution with low-latency data flow.
In addition, this results in a significant reduction in blur in the event of fast movement of objects or the camera.
Moreover, the high dynamic range resolution allows events to be recorded in both bright and dark regions.
This is especially important for outdoor solutions, where lighting is highly variable (e.g. in automotive: tunnel crossings, under overpasses, etc.).
From the perspective of designing energy-efficient solutions, these sensors seem particularly interesting: due to their asynchronous and sparse operation (recording only changes in brightness), they are on average more energy-efficient than typical digital cameras (frame cameras).

%Kamery zdarzeniowe rejestrują zmiany w natężeniu światła indywidualnie dla każdego piksela.
%Pojedyncze zdarzenie reprezentowane jest przez wektor $e = \{t, x, y, p\}$, gdzie $t$ to czas wystąpienia zdarzenia (timestamp), $x$ oraz $y$ współrzędne przestrzenne piksela rejestrującego zdarzenie, a $p$ przyjmuje wartość $-1$ lub $1$ w zależności kierunku zmiany.
%Akwizycja danych odbywa się więc tylko w momencie zmiany jasności piksela, co przekłada się na możliwość uzyskania danych o wysokiej rozdzielczości oraz z niską latencją.
%Dodatkowo powoduje to istotną redukcję zajwiska rozmycie w przypadku szybkiego ruchu -- obiektów lub kamery.
%Ponadto wysoka rozdzielczość dynamiczna (high dynamic range) pozwala na rejestrację światła zarówno w jasnych, jak i ciemnych rejonach, co ma duże znaczenie w kontekście rozwiązań zewnętrznych, gdzie oświetlenie jest bardzo zmienne (np. w automotive: przejazdy przez tunele, pod wiaduktami, itp.).
%Z perspektywy projektowania rozwiązań energooszczędnych czujniki te wydają się szczególnie interesujące: ze względu na nieciągłe działanie (rejestrację jedynie zmian jasności) są średnio bardziej wydajne energetycznie od kamer cyfrowych.

However, the format of event data (spatio-temporal sparse cloud) makes it impossible to use explicitly standard digital image processing algorithms, including those based on neural networks, for their analysis.
Therefore four possibilities are considered: designing a solution dedicated to event data (e.g. spiking neural networks -- SNN) \cite{b5}, processing a point cloud e.g. using graph neural networks (which can be challenging due to the different density of such a cloud in successive time segments) \cite{b6}, projecting (aggregating) events onto a plane and creating so-called pseudo-images \cite{b7}, or image reconstruction from such a cloud (which is computationally very complex) \cite{b8}.

%Jednakże, format danych zdarzeniowych (spatio-temporal sparse cloud) powoduje, że do ich analizy niemożliwe jest użycie wprost standardowych algorytmów przetwarzania obrazów cyfrowych, w tym również tych opartych o sieci neuronowe.
%Analiza takich danych przeprowadzana jest zatem jedną z czterech możliwości: projektowanie rozwiązania dedykowane danym zdarzeniowym (np. impulsowe sieci neuronowe), przetwarzanie chmury punktów np. wykorzystując grafowe sieci neuronowe (co może stanowić wyzwanie ze względu na różną gęstość takiej chmury w kolejnych odcinkach czasowych), rzutowanie (agregację) zdarzeń na płaszczyznę i tworzenie tzw. pseudo-obrazów lub rekonstrukcja obrazu z takiej chmury (która jest bardzo złożona obliczeniowo).

The use of pseudo-images best cooperates with the typical neural networks developed for images registered with frame cameras.
A number of such approaches have been proposed in the literature - they differ in the way they take into account the time dimension, in particular the change in signal over time.
In this work, we use the event frame method, which transforms events from a specific time slice (in our case 10ms) $\tau$ according to $f(u,t)=P_e(u)$ for $t - \sigma_e(u) \in (0; \tau)$ and $f(u,t)=0$ otherwise, where $\sigma_e$ denotes the time of the event $u(x,y)$ and $P_e$ its polarity.
This choice is dictated by two factors: firstly, the chosen object detection algorithm is dedicated to classical images, so the input format must be similar to them; secondly, due to the necessary limitations of memory and computational complexity for low-power devices, the method of converting the event cloud to a pseudo-image should not be too complex.
%Of course, a consequence of choosing this method is the omission of temporal information, but we assume that the generalisation properties of neural networks will compensate these shortcomings. 
Other pseudo-image representations worth mentioning are: simple binary frame \cite{b9}, exponentially decaying time surface \cite{b10} and event frequency \cite{b11}.

%Użycie pseudo-obrazów jest podejściem, które najlepiej łączy się z typowymi sieciami neuronowymi rozwijanymi dla obrazów z kamer klatkowych.
%W literaturze zaproponowano wiele tego typu podejść, które różnią się sposobem uwzględniania wymiaru czasu, w szczególności zmiany sygnału w czasie.
%W tej pracy wykorzystano metodę ramki zdarzeń (event frame), która przekształca zdarzenia z określonego odcinka czasowego (w naszym przypadku 10ms) $\tau$ zgodnie z $f(u,t)=P_e(u)$ dla $t - \sigma_e(u) \in (0; \tau)$ i $f(u,t)=0$ w pozostałych przypadkach, gdzie $\sigma_e$ oznacza czas zdarzenia $u(x,y)$, a $P_e$ jego polarność.
%Taki wybór podyktowany jest dwoma czynnikami: po pierwsze, wybrany algorytm detekcji obiektów jest dedykowany klasycznym obrazom, a zatem format wejścia musi być im zbliżony; po drugie, z uwagi na konieczne ograniczenia złożoności pamięciowo-obliczeniowej dla urządzeń niskiej mocy, metoda konwersji chmury zdarzeń do pseudo-obrazu nie powinna być zbyt skomplikowana.
%Oczywiście konsekwencją wyboru tej metody jest pominięcie informacji czasowych, zakładamy jednak, że właściwości generalizacji sieci neuronowych w pewien sposób zrekompensują te braki. 

In this research we used the Gen1 event dataset consisting of over 39 hours of recordings of urban, highway, suburbs and countryside scenes with a Prophesee Gen1 sensor attached to a car.
The collection was manually tagged based on greyscale images recorded simultaneously with the events, ultimately highlighting nearly $230000$ of cars and nearly $28000$ of pedestrians.
For the generation of pseudo-images, we used the tools provided by the authors of the database \cite{b7}. 
% Example images from the training set are shown in Fig. \ref{fig:dataset}.

%W niniejszej pracy wykorzystano zbiór danych zdarzeniowych Gen1 \cite{b25}, złożony z ponad 39 godzin nagrań urban, highway, suburbs and countryside scenes sensorem Prophesee Gen1 przytwierdzonym do samochodu.
%Zbiór został ręcznie oetykietowany na podstawie rejestrowanych równocześnie ze zdarzeniami obrazów w skali szarości, ostatecznie wyróżniając prawie $230000$ samochodów i prawie $28000$ pieszych.
%Do generacji pseudo-obrazów wykorzystano narzędzia udostępnione przez autorów bazy danych \cite{b7} - przykładowe obrazy ze zbioru uczącego pokazano na Rys. \ref{fig:dataset}.

% \begin{figure*}[!t]
% \centerline{\includegraphics[width=0.95\textwidth]{figures/dataset.png}}
%   \caption{Example pseudo-images generated using the event frame method from a point cloud recorded in a 10ms window.}
%   %\caption{Przykładowe pseudo-obrazy wygenerowane metodą ramki zdarzeń z chmury punktów rejestrowanych w oknie 10ms.}
%   \label{fig:dataset}
% \end{figure*}

% -----------------------------------------------------------------------------------------
\section{Related Work}
\label{sec:related_work}

In this section we present previous work on two topics closely related to our research. 
% In the first part we discuss methods for reducing the memory and computational complexity of neural networks, and the second part present previously published papers on object detection based on event data.

%W rodziale przedstawiono poprzednie prace dotyczące dwóch zagadnień ściśle związnaych z przedstawionymi badaniami. 
%W pierwszej cześci mówiono metody redukcji złożoności pamięciowej i obliczeniowej sieci neuronowych, a w drugiej dotychas opublikowane prace na temet detekcji obiektów na podstawie danych z kamery zdarzeniowej.

%%%%%%%%%%%%%%%%%%%%%%%%%%%%%%%%%%%%%%%%%%
\subsection{Reducing the memory and computational complexity of neural networks}

Standard methods for reducing the memory-computation complexity of neural networks include quantisation, pruning, and appropriate transformations in the form of layer fusion that do not change the inference result.
The first two methods lead to a reduction in the number of bits needed for parameter storage and the number of parameters, respectively.
In its simplest form, pruning, i.e. zeroing the connections or neurons with low impact on the final result, leads to a sparse network in which the number of arithmetic operations performed can be considerably reduced. 
However, this introduces the overhead of having to remember which weights or neurons have been zeroed (i.e. which calculations should be omitted).
An alternative is to train sparse networks, i.e. networks in which the structure of the removed connections is predetermined.
Since neural networks are generally redundant, appropriately performed pruning, or sparse network training, allows to achieve the performance of a dense network.

%Do standardowych metod redukcji złożoności pamięciowo-obliczeniowej sieci neuronowych zaliczamy kwantyzaję, pruning oraz odpowiednie transformacje w postaci fuzji warstw nie zmieniających wyniku inferencji.
%Dwie pierwsze metody prowadzą do redukcji kolejno liczby bitów potrzebnych do przechowywania parametrów oraz liczby parametrów.
%Pruning, czyli zerowanie połączeń lub neuronów o niskim wpływie na końcowy wynik w swojej najprostszej postaci prowadzi do otrzymania sieci rzadkiej, w której liczba wykonywanych operacji arytmetycznych może być zdecydowanie mniejsza, ale wprowadza narzut związany z koniecznością pamiętania, które wagi lub neurony zostały wyzerowane (tzn. które obliczenia należy pominąć).
%Alternatywą jest uczenie sieci rzadkich, czyli takich w których struktura wygaszonych połączeń jest z góry ustalona.
%Ponieważ sieci neuronowe są z reguły redundantne, odpowiednio przeprowadzony pruning, czy uczenie sieci rzadkiej pozwala na osiągnięcie skuteczności działania sieci gęstej.

For the same reason, a reduction in computational precision (quantisation) may also not result in a significant drop in performance relative to a floating-point network, while at the same time reducing memory complexity by several times.
A certain standard available in many tools, both software tools (e.g. tensorflow \cite{b12}, OpenVino \cite{b13}, brevitas \cite{b14}) but also in hardware implementations (e.g. FINN \cite{b15}, VitisAI \cite{b16}, nncf \cite{b17}) is linear quantisation, in particular to 8-bit integer values.

%Z tego samego powodu redukcja precyzji obliczeń (kwantyzacja) również może nie spowodować znacznego spadku skuteczności działania względem sieci zmiennoprzecinkowej, przy jednoczesnym kilkukrotnym zmniejszeniu złożoności pamięciowej.
%Pewnym standardem dostępnym w wielu narzędziach, zarówno software'owych (m.in. tensorflow, OpenVino, brevitas) ale również w implementacjach sprzętowych (m. in. FINN \cite{FINN}, VitisAI, nncf) jest kwantyzacja liniowa, w szczególności do wartości całkowitych ośmiobitowych.

This method reduces the model by a factor of four (in terms of number of bytes) and significantly simplifies the implementation of arithmetic operations, which is particularly important for low-power hardware implementations.
Linear quantisation to lower bit-widths can significantly affect the accuracy of the solution, therefore logarithmic quantisation methods, first proposed in \cite{b18}, are an interesting alternative.
This scheme allows to preserve the accuracy of the full-precision network for parameters 4-bit and smaller (after appropriate training of the quantised network) \cite{b1}, \cite{b2}.
In addition, this approach allows to replace the multiplication with a bit-shift, which, using the appropriate computing platform (GPU, eGPU, and especially reconfigurable platforms like FPGAs), results in reduced latency and reduced power consumption  \cite{b2}, \cite{b3}, \cite{b19}.
Obviously, the most radical form of reduction in computational precision is binary quantisation, however it is not without an impact on the accuracy of the network \cite{b20}, causing losses of several tens of per cent.

%Metoda ta pozwala na czterokrotne zmniejszenie modelu (w sensie liczby bajtów) oraz istotnie upraszcza realizację operacji arytmetycznych, co jest szczególnie istotne przy implementacjach sprzętowych niskiej mocy.
%Kwantyzacja liniowa do niższych szerokości bitowych może znacznie wpłynąć na skuteczność rozwiązania dlatego interesującą alternatywę stanowią metody kwantyzacji logarytmicznej, po raz pierwszy zaproponowane w \cite{b18}.
%Schemat ten pozwala na uzyskanie skuteczności sieci pełnej precyzji dla parametrów 4-bitowych i mniejszych (po odpowiednim uczeniu kwantyzowanej sieci) bez zmian w architekturze \cite{b1}, \cite{b2}.
%Dodatkowo podejście to pozwala zastąpić operację mnożenia przesunięciem bitowym, co przy wykorzystaniu odpowiedniej platformy obliczeniowej (GPU, eGPU, a w szczególności platformy rekonfigurowalne jak FPGA) skutkuje redukcją latencji oraz zmniejszeniem zużycia energii \cite{b2}, \cite{b3}, \cite{b19}.
%Oczywiście najbardziej radykalną formą redukcji precyzji obliczeń jest kwantyzacja binarna, natomiast nie pozostaje ona bez wpływu na skuteczność sieci \cite{b20}, powodując straty rzędu kilkudziesięciu procent.

Extremely low-precision networks, or rather new quantisation methods, are most often benchmarked for classification tasks, so it is difficult to say unequivocally whether the impact of different precision reduction methods is same for each task solved by a neural network, in particular for object detection.
However, it seems that due to the higher complexity and the need for higher precision information at the output, quantisation to very low bit-widths for detection task may have a greater impact on performance: especially if we consider the coordinates of the bounding box, rather than a class probability score. In principle, the latter should simply be higher for true class objects than for others, rather than with some specific value.
Several experiments with very low-precision quantisation of YOLO networks have been described in the literature.

In \cite{b21}, the authors note that quantisation of YOLO networks is challenging due to the problem of oscillation of weights between quantisation thresholds when learning low-precision networks, even at final epochs. 
%Sieci ekstremalnie niskiej precyzji, czy raczej nowe metody kwantyzacji, są najczęściej benchmarkowane dla zadań klasyfikacji, dlatego trudno jednoznacznie stwierdzić czy wpływ różnych metod redukcji precyzji obliczeń jest taki sam dla każdego zadania rozwiązywanego przez sieć neuronową, w szczególności dla detekcji.
%Wydaje się jednak, że ze względu na wyższy stopień skomplikowania i konieczność uzyskania informacji wyższej precyzji na wyjściu - tzn. współrzędnych prostokąta otaczającego, a nie miary prawdopodobieństwa przynależności do pewnej klasy, która z zasady powinna być po prostu wyższa dla obiektów klasy prawdziwej niż dla innych, a nie o pewnej konkretnej wartości - kwantyzacja do bardzo niskich szerokości bitowych może mieć większy wpływ na skuteczność niż w przypadku prostszych zadań.
%W literaturze opisano kilka eksperymentów z kwantyzacją bardzo niskiej precyzji sieci YOLO.
%W \cite{b21} autorzy zauważają, że kwantyzacja sieci YOLO jest wyzwaniem ze względu na problem oscylacji wag pomiędzy progami kwantyzacji w trakcie uczenia sieci niskiej precyzji, nawet przy końcowych epokach. 
The use of two additional methods EMA (Exponentially Moving Average) and QC (Quantisation Correction) have been proposed to minimise these oscillations ,in combination with quantised training. 
In the Quantisation Aware Training (QAT) version, EMA works similarly as in full precision network training, storing a history of both the weights and the scaling factors of the weights and activations.
With the introduction of EMA into QAT, the training is ﬁnitely smoother.
In addition, Quantisation Correction has been introduced after the training process, to minimise oscillations errors by linearly transforming the activations based on some pre-trained parameters.
The YOLOv5s network trained with such 4-bit quantisation on the COCO dataset achieves a mAP of $0.34$, with a baseline (full-precision network) mAP of $0.374$ -- thus a decrease of about $11\%$ (for more results for different architectures, we refer to the article).
In \cite{b22}, the authors proposed a Post-Training Quantisation method that achieved a mAP of $0.196$ for YOLOv5s and a COCO dataset for 4-bit weights.

%Zaproponowano użycie dwóch dodatkowych metod EMA (Exponentially Moving Average) oraz QC (Quantization Correction), które w połączeniu z uczeniem kwantyzowanym mają zminimalizować owe oscylacje. 
%W wersji QAT (Quantization Aware Training), EMA działa podobnie jak przy uczeniu sieci pełnej precyzji, przechowując historię zarówno wag, jak i współczynników skalowania wag i aktywacji.
%Dzięki wprowadzeniu EMA do QAT uczenie jest \textit{gładsze}.
%Dodatkowo wprowadzono Quantization Correction po procesie uczenia, która ma zniwelować błędy wynikające z oscylacji, za pomocą przekształcenia liniowego aktywacji na podstawie pewnych wyuczonych wcześniej parametrów.
%Tak nauczona sieć YOLOv5s przy kwantyzacji 4-bitowej dla zbioru COCO osiąga mAP $0.34$, przy mAP $0.374$ dla sieci pełnej precyzji - a zatem spadek o ok. $11\%$ (po więcej wyników dla różnych architektur odsyłamy do artykułu).
%W \cite{b22} autorzy zaproponowali Post-Training Quantization metodę, która pozwoliła na osiągnięcie mAP $0.196$ dla YOLOv5s i zbioru COCO dla 4-bitowych wag.

It is worth noting that both of these methods propose the use of a linear quantisation scheme with additional mechanisms to prevent significant drops in efficiency, as achieved in \cite{b21}. 
In this paper, we propose a mixed precision solution, with the network weights quantised logarithmically to powers-of-two, and the rest of the activation and parameters quantised linearly.

Such a method allows a significant reduction in computational complexity due, firstly, to the possibility of changing the multiplication operation to a bit shift and, secondly, performing the remaining computations as integer operations.

%Warto podkreślić, że obie te metody proponują użycie schematu kwantyzacji liniowej z dodatkowymi mechanizmami, które mają zapobiec znacznym spadkom skuteczności, tak jak to osiągnięto w \cite{b21}. 
%W niniejszej pracy proponujemy rozwiązanie mieszanej precyzji, o wagach sieci kwantyzowanych logarytmicznie do potęg dwójki i reszcie aktywacji i parametrów kwantyzowanych liniowo.

%Taka metoda pozwala na znaczą redukcję złożoności obliczeniowej ze względu po pierwsze na możliwość zmiany operacji mnożenia na przesunięcie bitowe, a po drugie realizację pozostałych operacji na liczbach całkowitoliczbowych.

The gains from converting multiplication operations to bit shifting have been described in several papers.
In \cite{b1}, the authors implemented GPU kernels for PoT-weighted computing and showed gains over standard filtering with multiplication, shortening the inference time for ResNet18 by $25\%$.
In \cite{b23} the authors show gains in energy and logic resources used for 5-bit logarithmic quantisation relative to standard 16-bit multiplier for UMC 55nm Low Power Process.
The paper \cite{b24} presents dynamic power consumption for a hardware implementation of matrix multiplication for uniformly and PoT quantised values. At 3 bits, one can assume a decrease in power requirements of about $20\%$, at 5 bits -- $35\%$, in favour of PoT weights.
As was previously shown in  \cite{b19}, the hardware convolution layer accelerator on the Zynq UltraScale+ MPSoC ZCU104 development board with a SoC FPGA chip, with 4-bit PoT weights can be at least $1.4x$ more power efficient than the linearly quantised version, and the difference in power efficiency increases at higher chip clock frequencies.

%Zyski płynące ze zamiany operacji mnożenia na przesunięcie bitowe zostały opisane w kilku pracach.
%W \cite{b1} autorzy zaimplementowali kernele GPU do obliczeń z uwzględnieniem wag PoT i pokazali zyski względem standardowej filtracji z mnożeniem, skracjąc czas inferencji dla sieci ResNet18 o $25\%$.
%W \cite{b23} pokazują zyski energetyczne oraz użytych zasobów elektronicznych dla 5-bitowej kwantyzacji logarytmicznej względem standardowej 16 bitowej mnożarki dla UMC 55nm Low Power Process.
%W pracy \cite{b24} zaprezentowano pobór dynamic power dla sprzętowej implementacji mnożenia macierzowego dla wartości uniformly and PoT quantized. Przy 3 bitach można przyjąć spadek w zapotrzebowaniu na moc ok. $20\%$, przy 5 bitach -- $35\%$, na korzyść wag PoT.
%Jak wcześniej wykazaliśmy w  \cite{b19}, sprzętowy akcelerator warstw konwolucyjnych w układzie FPGA platformy Zynq UltraScale+ MPSoC ZCU104, z 4-bitowymi wagami PoT może być przynajmniej $1.4x$ bardziej wydajny energetycznie od wersji z kwantyzowanej liniowo, a różnica w wydajności energetycznej wzrasta przy wyższych częstotliwościach taktowania układu.

In this paper, we propose a mixed-precision detector that optimises all parameters of a neural network: to our knowledge, all previous work on logarithmic quantisation has referred to bit-width reduction mainly of weights and, less often, also of activation, neglecting the issue of the batch normalisation layer.
Although this layer is not a necessary part of building neural network architectures, it is a certain standard in most classical solutions.
It is therefore important to propose a method that is usable for virtually any ready-made full-precision model.
%Furthermore, we also demonstrate the performance gains of implementing a mixed precision model, with mixed quantisation and bit-shifting, relative to using a linear-only quantised model in the well-known FINN framework.

%W niniejszej pracy proponujemy detektor mieszanej precyzji, który optymalizuje wszystkie parametry sieci neuronowej: według naszej wiedzy, wszystkie poprzednie prace o kwantyzacji logarytmicznej odnosiły się do redukcji szerokości bitowej głównie wag i rzadziej również aktywacji, pomijając kwestię warstwy normalizacji partiami.
%Chociaż warstwa ta nie jest niezbędnym elementem budowania architektur sieci neuronowych, jest pewnym standardem w większości klasycznych rozwiązań.
%Istotne zatem jest zaproponowanie metody, która jest możliwa do użycia dla właściwie każdego gotowego modelu pełnej precyzji.
%Ponadto, demonstrujemy też zyski wydajnościowe płynące z implementacji modelu mieszanej precyzji, z mieszaną kwantyzacją i przesunięciem bitowym, względem użycia modelu kwantyzowanego jedynie liniowo, w znanym frameworku FINN.

%%%%%%%%%%%%%%%%%%%%%%%%%%%%%%%%%%%%%%%%%%
\subsection{Object detection based on event data}

Work on pedestrian and vehicle detection on the GEN1 event dataset \cite{b25} can be found in the literature.
The dataset itself was created from more than 39 hours of recordings using the GEN1 event sensor from the Prophesee company, collecting data at a resolution of 304x240.
% The authors then manually labelled the data, resulting in more than 255,000 labels -- bounding boxes together with a class label: vehicle (228,123 labels) or pedestrian (27,658).

%W literaturze można znaleźć prace poświęcone detekcji pieszych i pojazdów na zbiorze danych zdarzeniowych GEN1 \cite{b25}.
%Sam zbiór powstał na podstawie ponad 39 godzin nagrań z wykorzystaniem sensora zdarzeniowego GEN1, zbierającego dane z rozdzielczością 304x240.
%Następnie autorzy ręcznie etykietowali dane, w efekcie uzyskując ponad 255000 etykiet - prostokątów otaczających wraz z etykietą klasy: pojazdu (228123 etykiet) lub pieszego (27658).
% Sieci impulsowe (i podobne). 

In \cite{b26} authors proposed general methods for converting models trained on image-like event representations to models that operate asynchronously.
Event data are represented by a Sparse Recursive Representation, which can be updated as new events arrive in a sparse manner (i.e. only where a change has occurred).
This form of input allows the use of so-called sparse convolutions, recalculating only those activations which value may have changed.
Using such a model, a mAP of $0.129$ was obtained for the GEN1 set and event histogram representation ($0.145$ using a standard convolutional neural network).

%W \cite{b26} zaproponowano ogólne metody konwersji modeli uczonych na image-like event representations na modele działające asynchronicznie.
%Dane zdarzeniowe są reprezentowane przez reprezentację Sparse Recursive Representation, która może być aktualizowana w miarę przychodzenia nowych zdarzeń w sposób rzadki (tzn. jedynie tam, gdzie nastąpiła zmiana).
%Taka postać wejścia umożliwia wykorzystanie tzw. rzadkich konwolucji, przeliczając tylko te aktywacje, których wartość mogła się zmienić.
%Za pomocą takiego modelu uzyskano mAP $0.129$ dla zbioru GEN1 i reprezentacji event histogram ($0.145$ przy użyciu standardowej konwolucyjnej sieci neuronowej).
The paper \cite{b27} proposes a directly trained spiking network -- EMS-YOLO.
The first convolution layer is trained by converting the input to spikes, the subsequent layers consist of EMS-Modules, which are fully spiking residual blocks with LIF (Leaky Integrate and Fire) activation function.
The input of the network is an image-based 2D representation of events within a specific time window.
On the GEN1 set, mAP of $0.267$ was achieved, for inference over 4 time steps, with a Firing Rate of $21.15\%$ neurons, and a network with $6.20M$ parameters.

%W pracy \cite{b27} zaproponowano uczoną bezpośrednią impulsową sieć EMS-YOLO, w której pierwsza warstwa konwolucyjna uczona jest konwersji wejścia do impulsów, na kolejne składają się bloki EMS (EMS-Modules), czyli w pełni impulsowe bloki rezydualne z funkcją aktywacji LIF.
%Wejście sieci stanowi obrazowa reprezentacja 2D zdarzeń w określonym oknie czasowym.
%Na zbiorze GEN1 uzyskano skuteczność mAP $0.267$ dla inferencji przez 4 kroki czasowe, przy częstotliwości zapalania (Firing Rate) neuronów $21.15\%$ i sieci z $6.20M$ parametrów.

A spiking feature pyramid network (SpikeFPN), consisting of an encoder backbone spiking neural network, a spiking feature pyramid building network and a spiking multi-head-detection module, was presented in \cite{b5}.
Event data are represented using Stacking Based on Time (SBT), which, with standard compression of the event stream into a frame, allows some temporal information to be retained.
A mAP of $0.223$ was achieved for the GEN1 dataset.

%W \cite{b5} przedstawiono spiking feature pyramid network (SpikeFPN), składającej się z encoder backbone spiking neural network, impulsowej sieci budującej piramidę cech oraz impulsowego modułu multi-head-detection.
%Dane zdarzeniowe reprezentowane są z użyciem metody Stacking Based on Time (SBT), która przy standardowej %kompresji strumienia zdarzeń do ramki pozwala na zachowanie pewnych informacji czasowych.
%Dla zbioru GEN1 uzyskano mAP $0.223$.

The Group Event Transformer architecture is proposed in \cite{b28}, which is dedicated to process event data directly and thus without the need to convert events to a 2D representation, which results in the loss of some timing or polarity information.
Events are embedded into Group Tokens based on timing and polarity information, and then processed by a transformer-based architecture using self-attention mechanisms.
On the GEN1 dataset, the proposed model achieves mAP of $0.406$, and with additional memory mechanisms to support the challenges of capturing only moving objects by the event camera, $0.484$.

%W \cite{b28} zaproponowano architekturę Group Event Transformer, która dedykowana jest danym w postaci zdarzeniowej bezpośrednio (a więc bez konieczności konwersji zdarzeń do \textit{jakiejś} reprezentacji 2D, co powoduje utratę pewnych informacji czasu i/lub biegunowości).
% Zdarzenia są embedowane w Group Tokens na podstawie informacji czasowej i biegunowości, a następnie przetwarzane przez architekturę opartą o transformersy z wykorzystaniem mechanizmów self-attention.
% Na zbiorze GEN1 zaproponowany model osiąga skuteczność mAP $0.406$, a z dodatkowymi mechanizmami pamięci wspomagającymi wyzwania związane z rejestracją obiektów jedynie ruchomych przez kamerę zdarzeniową, $0.484$.

Similarly, in \cite{b29} the authors also propose to analyse events directly, in the form of a structured (in terms of spatial dimensions) point cloud.
At each time step, the model based on the Vision Transformer architecture processes a new data stream together with previous states from recursive layers.
The recursive layers help to solve the problem of slow-moving objects not captured by the event camera at each time step.
The described model achieves mAP of $0.472$ on the GEN1 dataset, for the largest of the proposed architectures (for the 4x smaller model mAP $0.441$).

%Podobnie w \cite{b29} autorzy również proponują analizę bezpośrednio zdarzeń, w postaci ustrukturyzowanej (w sensie wymiarów przestrzennych) chmury punktów.
%W każdym kroku czasowym model oparty o architekturę Vision Transformers przetwarza nowy strumień danych wraz z poprzednimi stanami z warstw rekurencyjnych.
%Warstwy rekurencyjne pomagają rozwiązywać problem obiektów wolno poruszających się, nie rejestrowanych przez kamerę zdarzeniową w każdym odcinku czasowym.
%Opisany model osiąga mAP na poziomie $0.472$ na zbiorze GEN1, dla największej z proponowanych architektur (dla modelu 4x mniejszego mAP $0.441$).

A recursive architecture consisting of standard convolution layers and LSTM blocks is presented in \cite{b7}. 
The model directly processes the event data stream as a tensor with fixed spatial dimensions.
The proposed method achieves mAP of $0.4$ for GEN1 dataset, with $24.1M$ parameters.
In \cite{b30} the authors proposed to use the YOLO (You Only Look Once) model, together with an adaptive conversion of the events to Hyper Histogram form, which allows the preservation of both temporal and polarity information.
In addition, a corresponding modification of one of the key data augmentation methods for YOLO detectors in the form of Shadow Mosaic was also proposed.
The model trained in such a way achieves mAP of $0.47$, while the analogous solution without Hyper Histogram and Shadow Mosaic $0.394$ (treated by the authors as a baseline).
In \cite{b31} the authors proposed another way of representing event data - Temporal Active Focus (TAF) - along with a small Agile Event Detector (AED) neural network model with a corresponding event encoder extracting semantic information from temporal to flat vector input.
The architecture of the detector itself is based on Darknet21 and head like for YOLOX.
Ultimately, the model achieves mAP of $0.454$, with $14.8M$ of parameters and a baseline (YOLOX architecture and Event Volume representation) of $0.35$.

% Sieci raczej zwykłe
%W \cite{b7} przedstawiono rekurencyjną architekturę złożoną ze standardowych warstw konwolucyjnych i bloków LSTM. 
%Model przetwarza bezpośrednio strumień danych zdarzeniowych w postaci tensora o ustalonych wymiarach przestrzennych.
%Zaproponowana metoda pozwala na osiągnięcie mAP $0.4$ dla zbioru GEN1 z $24.1M$ parametrów.
%W \cite{b30} zaproponowano użycie modelu Yolo wraz z adaptacyjną konwersją zdarzeń do postaci Hyper Histogram, która pozwala na zachowania informacji zarówno czasowych jak i biegunowości.
%Dodatkowo zaproponowano też odpowiednią modyfikację jednej z kluczowych metod augmentacji danych dla detektorów Yolo w postaci Shadow Mosaic.
%Tak uczony model osiąga mAP $0.47$, natomiast analogiczne rozwiązanie bez Hyper Histogram oraz Shadow Mosaic $0.394$ (traktowane przez autorów jako baseline).
%W \cite{b31} zaproponowano kolejny sposób reprezentacji danych zdarzeniowych Temporal Active Focus (TAF) i niewielki model sieci neuronowej Agile Event Detector (AED) z odpowiednim enkoderem zdarzeń wyciągającym informacje semantyczne z czasowych do postaci płaskiego wektora na wejściu. 
%Architektura samego detektora bazuje na sieci Darknet21 i head jak dla YOLOX.
%Ostatecznie model osiąga mAP $0.454$ z $14.8M$ parametrów przy baseline (architekturze YOLOX i reprezentacji Event Volume) $0.35$.

Table \ref{tab:prev_work} summarises the discussed previous work on object detection on the GEN1 dataset.
It is worth noting that these solutions are classical in the sense of computational precision, i.e. they operate on 32- or 16-bit floating point numbers, and mostly require GPU acceleration.
For solutions based on spiking neural networks (\cite{b26}, \cite{b27}, \cite{b5}), the firing rate can be as low as $20\%$, which of course translates into lower computational complexity, but these solutions have much lower accuracy (and the memory complexity remains constant).
Independently, due to their size, efficient use of the mentioned models in low-power devices directly is virtually impossible.

%W tabeli \ref{tab:prev_work} zestawiono omawiane wcześniejsze prace dotyczące detekcji obiektów na zbiorze GEN1.
%Warto podkreślić, że są to rozwiązania klasyczne w rozumieniu precyzji obliczeń, tj. operują na liczbach zmiennoprzecinkowych 32- lub 16-bitowych, i w większości wymagają akceleracji z użyciem GPU.
%Dla rozwiązań opartych o impulsowe sieci neuronowe (\cite{b26}, \cite{b27}, \cite{b5}) firing rate może być na poziomie nawet $20\%$, co oczywiście przekłada się na niższą złożoność obliczeniową, ale rozwiązania te charakteryzuje znacznie niższa skuteczność (a złożoność pamięciowa pozostaje niezmienna).
%Niezależnie, ze względu na ich wielkość, efektywne użycie wymienionych modeli w urządzeniach niskiej mocy wprost jest właściwie niemożliwe.
\begin{table}[!t]
\small
\caption{Overview of existing object detectors for the GEN1 dataset, distinguishing mAP and number of model parameters.} 
%\caption{Przegląd istniejących detektorów dla zbioru GEN1 z wyróżnieniem mAP oraz liczby parametrów modelu.
\label{tab:prev_work}
\newcolumntype{C}{>{\centering\arraybackslash}X}
\begin{tabularx}{0.49\textwidth}{CCC}
\toprule
Method & mAP & Model size \\
\midrule
ASCNN \cite{b26} & $0.129$ & $133 $M \\
EMS-ResNet \cite{b27} & $0.267$ & $6.2 $M \\
SpikeFPN \cite{b5} & $0.223$ & $11 $M \\
GET \cite{b28} & $0.479$ & $21.9 $M \\
RVT-S \cite{b29} & $0.465$ & $9.9 $M \\
RVT-T \cite{b29} & $0.441$ & $4.4 $M \\
Gray-RetinaNet \cite{b7} & $0.44$ & $32.8 $M \\
YOLOv5l \cite{b30} & $0.47$ & $46.5 $M \\
AED \cite{b31} & $0.454$ & $14.8 $M \\
\bottomrule
\end{tabularx}
\end{table}

For this article, the last two methods proposed by \cite{b30} and \cite{b31} are particularly relevant: the former because of the used YOLO architecture, and the latter because of the small size of the network.
To the authors' best knowledge, these models determine the current SoTA (in the sense of highest detection accuracy) on the GEN1 dataset, and it is to these models that we will compare our proposed mixed precision network.
The analysis shows that, besides the choice of architecture (spiking network, classical deep detectors or transformers), the representation of the input data itself (multidimensional tensor or 2D image), but also its augmentation, is important.

%Z punktu widzenia niniejszego artykułu szczególnie istotnymi metodami dwie ostatnie, proponowane przez \cite{b30} oraz \cite{b31}: pierwsza ze względu na użytą architekturę YOLO, druga natomiast ze względu na niewielki rozmiar sieci.
%Według wiedzy autorów modele te wyznaczają aktualne SoTA (w sensie najwyższej skuteczności detekcji) na zbiorze GEN1 i to do tych modeli będziemy porównywać proponowaną przez nas sieć mieszanej precyzji.
%Z przeprowadzonej analizy wynika, że obok wyboru architektury (sieci impulsowej, klasycznych głębokich detektorów czy transformersów) istotna jest sama reprezentacja danych wejściowych (wielowymiarowego tensora lub obrazu 2D), ale również ich augmentacja.

In addition, object detection on event frames faces the problem of continuity of information in case of objects moving at variable speeds (in particular low or even static). 
This is an issue not encountered in such form for algorithms designed for standard images.
However, this paper does not aim to make further improvements in these areas, but to propose a method with the lowest possible memory-computational complexity, while maintaining the highest performance.
Reducing memory-computational complexity promotes efficient information processing in embedded devices, particularly those with the potential for significant parallelisation of computation: it can provide low latency while requiring little energy.

%Dodatkowo detekcja obiektów na ramkach zdarzeniowych mierzy się z problemem ciągłości informacji w przypadku obiektów poruszających się ze zmienną prędkością (w szczególności niską lub wręcz statycznych). 
%Jest to zagadnienie niespotykane w tej formie dla algorytmów działających na klasycznych obrazach.
%Niniejsza praca nie ma jednak na celu wprowadzenia kolejnych usprawnień w tych obszarach, a zaproponowanie metody o możliwie najniższej złożoności pamięciowo-obliczeniowej, przy jednoczesnym zachowaniu najwyższej skuteczności działania.
%Redukcja złożoności pamięciowo-obliczeniowej sprzyja efektywnemu przetwarzaniu informacji w urządzeniach wbudowanych, w szczególności tych o możliwościach znacznej równoleglizacji obliczeń: może zapewnić niską latencję przy jednoczesnym małym zapotrzebowaniu energetycznym.

%%%%%%%%%%%%%%%%%%%%%%%%%%%%%%%%%%%%%%%%%%
\section{PowerYOLO Network}
\label{sec:eyolo}

One of the most popular algorithms for object detection are networks based on the YOLO (You Only Look Once) architecture \cite{b32}.
However, due to their memory and computational complexity, proper optimisation is required for efficient inference of such architectures on low-power devices.
In this section, we first discuss the complexity of the YOLO network and then present the proposed mixed-precision quantisation -- power-of-two and linear.

%Jedną z bardziej popularnych metod detekcji obiektów są sieci oparte o architekturę YOLO (You Only Look Once) \todo{cite}.
%Ze względu na swoją złożoność pamięciowo-obliczeniową, w celu wydajnej inferencji takich architektur w urządzeniach niskiej mocy konieczna jest jednak ich odpowiednia optymalizacja.
%W rozdziale najpierw przedawiamy analizę złożoności sieci YOLO, a następnie prezntujemy proponowaną mixed-precsision quantisation - power-of-two i linear.

\subsection{YOLO computational and memory complexity}
\label{subsec:yolo}
% Sprawdzone PL

The neural networks from the YOLO ``family'' consist of three main blocks: backbone for image feature extraction, neck for feature pyramid extraction (by combining features from different levels of the backbone) and head as the final stage determining bounding boxes, class labels, probabilities and objectness scores.
The architecture of the YOLO network has evolved over the years, and the latest version widely accepted by the community as SoTA is YOLOv8, available open source at \cite{b32}.
Depending on the number of backbone layers, the network comes in several sizes (by convention: n, s, m, l, x), and for the purpose of this paper the commonly used (due to its good ratio of memory and computational complexity to achieved accuracy) YOLOv8s version was chosen.

%Sieci neuronowe z rodziny YOLO składają się z trzech głównych bloków: backbone do ekstrakcji cech obrazu, neck do ekstrakcji piramid cech (poprzez kombinację cech z różnych poziomów bakcbone) oraz head jako ostatni etap wyznaczający prostokąty otaczające, wskazania klas oraz objectness score.
%Architektura sieci YOLO ewoluowała na przestrzeni lat, a najnowsza wersja powszechnie zaakceptowana przez community jako SoTA to YOLOv8, dostępna otwartoźródłowo w \cite{b32}.
%W zależności od liczby warstw backbone, sieć występuje w kilku rozmiarach (zwyaczajowo: n, s, m, l, x), a na potrzeby niniejszej pracy wybrano powszechnie używaną (ze względu na dobry stosunek złożoności pamięciowo-obliczeniowej do osiąganej skuteczności) wersję YOLOv8s.

% The selected network consists of 214 layers, including 64 convolution layers.
The entire network has 7.2 M parameters (and thus, in the classical approach, 32- or 16-bit floating-point values), of which almost $79\%$ are the weights of the convolution layers.
Linear quantisation of the weights and activations to 8-bit numbers generally preserves the accuracy of the full-precision network, while reducing the memory complexity by 4x (relative to 32-bit numbers) and increasing the throughput of the mathematical operations by 16x \cite{b33}.
If a further reduction in the precision of the convolution layer weights to 4-bit values is made, a very compact architecture both in terms of memory and computation is obtained. 
In such a case, according to \cite{b33}, the throughput of mathematical operations on the GPU increases up to x32.
Implementing such a mixed-precision network (with 4-bit weights and 8-bit activations) in both eGPUs and dedicated processors therefore allows for significant improvements in inference time, while reducing memory complexity by almost x8. 
Using a logarithmic quantisation scheme of weights to powers of two, we will additionally enable the conversion of multiply-accumulate (MAC -- Multiply ACcumulate) operations to shift-accumulate (BAC -- Bitshift ACcumulate) operations, ultimately introducing a radical simplification in neural network computation.

%Wybrana sieć składa się z 214 warstw, w tym 64 konwolucyjnych.
%Cała sieć posiada 7.2 M parametrów (a zatem w klasycznym podejściu wartości zmiennoprzecinkowych 32 lub 16 bitowych), z czego prawie $79\%$ to wagi warstw konwolucyjnych.
%Liniowa kwantyzacja wag i aktywacji do liczb 8-bitowych z reguły pozwala na zachowanie skuteczności działania sieci pełnej precyzji, przy jednoczesnej redukcji złożoności pamięciowej 4x (względem liczb 32-bitowych) i zwiększenie przepustowości operacji matematycznych 16x \cite{b33}.
%Jeżeli dokonać dalszej redukcji precyzji wag warstw konwolucyjnych do wartości 4-bitowych, otrzymamy bardzo kompaktową architekturę, zarówno pod kątem pamięciowym i obliczeniowym -- w takim przypadku, zgodnie z \cite{b33}, przepustowość operacji matematycznych na GPU wzrasta nawet x32.
%Implementacja takiej sieci mieszanej precyzji (z 4-bitowymi wagami i 8-bitowymi aktywacjami) zarówno w eGPU, jak i dedykowanych procesorach pozwala zatem na znaczną poprawę czasu inferencji, przy jednoczesnej redukcji złożoności pamięciowej niemal x8. 
%Korzystając ze schematu kwantyzacji logarytmicznej wag do wartości potęg dwójki, dodatkowo umożliwimy zamianę operacji mnożąco-akumulujących (MAC - Multiply ACcumulate) na przesuwno-akumulujące (BAC - Bitshift ACcumulate), co ostatecznie wprowadzi radykalne uproszczenia w obliczeniach sieci neuronowej.

% ---------------------------------------------------------------------------
\subsection{Mixed Precision Quantization}
\label{subsec:mixedq}

In this work, we propose to use mixed quantisation, with convolution layer weights quantised logarithmically to 4-bit powers of two (so each weight can be represented by $2^n$ where $n$ is integer) and activations quantised linearly to 8-bit values.
Bias is customarily quantised to a 32-bit number to avoid overflow during the accumulation of convolution results.
Linear quantisation of 8-bit activations allows the network to maintain high performance, without the need to re-train the quantised network, but only by calibrating the values of the scaling factors and activation offset on some representative set of input data (customarily a subset of the training data).

% Sprawdzone PL
%W niniejszej pracy proponujemy użycie kwantyzacji mieszanej, z wagami warstw konwolucyjnych kwantyzowanych logarytmicznie do 4-bitowych wartości potęg dwójki oraz aktywacji kwantyzowanych liniowo do wartości 8-bitowych.
%Bias zwyczajowo kwantyzowany jest do liczby 32-bitowej, aby nie dopuścić do overflow w trakcie akumulacji wyników konwolucji.
%Kwantyzacja liniowa 8-bitowa aktywacji pozwala na utrzymanie wysokiej skuteczności sieci, bez konieczności uczenia sieci kwantyzowanej, a jedynie dokonując kalibracji wartości współczynników skalującego i offsetu aktywacji na pewnym reprezentatywnym zbiorze danych wejściowych (zwyczajowo podzbioru danych uczących).

% At the same time, linear quantisation does not guarantee to maintain the highest efficiency when using lower bit-widths.
Using PoT weights quantisation, which concentrates more quantisation intervals around zero thus mimics the shape of the distribution of weight values in the convolution layer, we can achieve better accuracy results with lower than standard 8-bit bit widths.% (Fig. \ref{fig:pot_mot}).

PoT quantisation is used in the so-called quatisation aware training (QAT).
The weights of the quantised network are initialised with the trained full-precision network and then the quantised network is trained (fine-tuned) for a relatively small number of epochs.
Due to the discontinuity of the quantisation function, the gradient method cannot be used explicitly.
The forward pass is done using quantised weights, and during backpropagation the gradients are computed using full precision equivalents (after the weights are updated, the network is re-quantised).

%Kwantyzacja (Eq. \ref{eq:loq_quant}) używana jest w tak zwanym procesie uczenia kwantyzowanego (QAT).
%Wagi sieci kwantyzowanej inicjalizowane są nauczoną siecią pełnej precyzji, a następnie przez stosunkowo niewielką liczbę epok uczy się sieć kwantyzowaną.
%Ze względu na nieciągłość funkcji kwantyzacji, metoda gradientowa nie może być użyta wprost, zatem przejście w przód odbywa się z użyciem wag kwantyzowanych, a gradienty obliczane są z użyciem odpowiedników pełnej precyzji (po aktualizacji wag sieć jest ponownie kwantyzowana).

Another method of reducing memory and computational complexity in neural network inference is the fusion of convolution layers with batch normalisation layers according to Eq.\eqref{eq:bn_fusion_1}:
\begin{equation} \label{eq:bn_fusion_1}
\small
    y_{bn} = \frac{x_{conv}-\mu}{\sqrt{\sigma^2 - \epsilon}} \gamma + \beta = \frac{\gamma}{\sqrt{\sigma^2 - \epsilon}} x_{conv} - \mu \frac{\gamma}{\sqrt{\sigma^2 - \epsilon}} + \beta
\end{equation}
where $x_{conv}$ is the output of the convolution layer. Thus, the weights $w$ and the bias $b$ of the filters for inference are modified accordingly: $w_{fused} = \frac{\gamma}{\sqrt{\sigma^2 - \epsilon}} * w$ and $b_{fused} = b - \mu \frac{\gamma}{\sqrt{\sigma^2 - \epsilon}} + \beta $.

Naturally, such a modification does not affect the accuracy of the neural network, but it reduces the required number of multiplication and addition operations, and the number of parameters.
Introducing exactly such a fusion to a power-of-two network would result in the weights no longer being in the form of powers of two, and therefore losing an important property of the described method.
In order to reduce the number of operations, we propose a small yet powerful modification.
%to introduce two different procedures leading to identical simplifications.
First, since the bias is quantised linearly, it can be successfully modified according to the standard fusion scheme (Eq. \eqref{eq:bn_fusion_1}).
Next, we propose to combine the multiplier usually associated with weights not with weights, but with the scaling factor of the quantisation operation.
In this way, we reduce as much as possible all the additional computations introduced by batch normalisation layer, while only slightly increasing the memory complexity relative to the model after standard fusion - instead of one scaling factor for the whole layer, we get scaling factors for each output feature map separately, and the batch normalisation completes during re-quantisation between layers.

Ultimately, we obtain a mixed-precision model, efficient in terms of memory and computational complexity, for which the calculations in the standard convolution layer -- batch normalisation block can be written with the simplified equation \eqref{eq:quant_flow}, where $\phi = \sqrt{\sigma^2 - \epsilon}$:
\begin{equation} \label{eq:quant_flow}
\small
y_{q} = \frac{\gamma}{\phi} \frac{s_x s_w}{s_y} (\sum_{i=0}^{k} \sum_{j=0}^{k} x_q^{ij} >> w_{pot}^{ij} + B_q)
\end{equation}
where $s_w, s_x, s_y$  are the scaling factors for quantising the weights, block inputs and block outputs, $k$ is the dimensions of the convolution filter, $x_q$is the activation from the previous layer quantised to INT8, $B_q$ is the modified bias $B = \frac{\gamma(b - \mu)}{\phi} + \beta$ quantised to INT16, and $w_{pot}$ are weights in the INT4 format, in powers of two.

%Ostatecznie otrzymujemy więc wydajny w sensie złożoności pamięciowo-obliczeniowej model mieszanej precyzji, dla którego obliczenia w standardowym bloku warstw konwolucyjnej - normalizacji partiami można zapisać uproszczonym równaniem \eqref{eq:quant_flow}, gdzie $\phi = \sqrt{\sigma^2 - \epsilon}$:
%\begin{equation} \label{eq:quant_flow}
%y_{q} = \frac{\gamma}{\phi} \frac{s_x s_w}{s_y} (\sum_{i=0}^{k} \sum_{j=0}^{k} x_q^{ij} >> w_{pot}^{ij} + B_q)
%\end{equation}
%gdzie $s_w, s_x, s_y$ to współczynniki skalujące dla kwantyzacji kolejno wag, wejścia i wyjścia bloku, $k$ to wymiary filtra konwolucyjnego, $x_q$ to aktywacja z poprzedniej warstwy kwantyzowana do INT8, $B_q$ to zmodyfikowany bias $B = \frac{\gamma(b - \mu)}{\phi} + \beta$ kwantyzowany do INT16, a $w_{pot}$ to wagi w formacie INT4, w postaci potęg dwójki.

For re-quantisation between successive layers of the neural network during inference, for PoT quantisation the factor $\frac{\gamma}{\phi} \frac{s_x s_w}{s_y}$ is in a form of vector. 
In comparison, in case of linear quantisation with standard convolution and batch normalisation fusion, it's a scalar (with value $\frac{s_x s_w}{s_y}$).
For simplicity, the indices indicating the output feature map are omitted in the equation.

%Z punktu widzenia rekwantyzacji między kolejnymi warstwami sieci neuronowej w trakcie inferencji istotny jest współczynnik $\frac{\gamma}{\phi} \frac{s_x s_w}{s_y}$ będący w przypadku kwantyzacji PoT wektorem, w przeciwieństwie do przypadku kwantyzacji liniowej, ze standardową fuzją warstw konwolucyjnej i normalizacji partiami (i przyjmujący wtedy wartość skalarną $\frac{s_x s_w}{s_y}$).
%Dla uproszczenia w równaniu pominięto indeksy wskazujące wyjściową mapę cech.

Although the granularity of quantisation can be chosen in many ways, in this paper we chose the layer-wise quantisation.
This decision is based on the following facts: firstly, the shape of the distribution of weights in the convolution layer of the network is logarithmic; secondly, other schemes are characterised by higher computational complexity.
For example in the case of the channel-wise scheme, quantisation is performed for each filter separately, which significantly increases the training time. 
For the network in question, experiments indicated as much as a 25 times increase in the duration of one epoch.
However, due to the proposed convolution and batch normalisation fusion scheme, channel-wise quantisation would not introduce additional computation during inference, and the impact on accuracy remains an open topic.
Nonetheless, as shown in section \ref{sec:results}, the layer-wise scheme allows for very high accuracy.

\section{Results}
\label{sec:results}

In this section, we present the results of the proposed PowerYOLO network for the task of object detection on event camera GEN1 dataset. We present comparison both with other solutions for the same task, and with other networks belonging to the TinyML group. In addition, we show the applicability of the obtained results for building a hardware accelerator.

%W rozdziale przedstawmy wyniki propowanej sieci PowerYOLO dla zadania detekcji obiektów na podstawie danych z kamery zdarzeniowej. Porównujemy je zarówno z innymi podobnymi rozwiązanianimi, jak i z innymi sieciami zaliczanymi do nurtu TinyML. Poandto pokazujemy możlwiości zastosowania uzyskanych wyników przy budowanie akceleratora sprzętowego.

\subsection{Mixed Precision YOLO}
% Sprawdzone PL

\begin{figure}[!t]
\centerline{\includegraphics[width=0.35\textwidth]{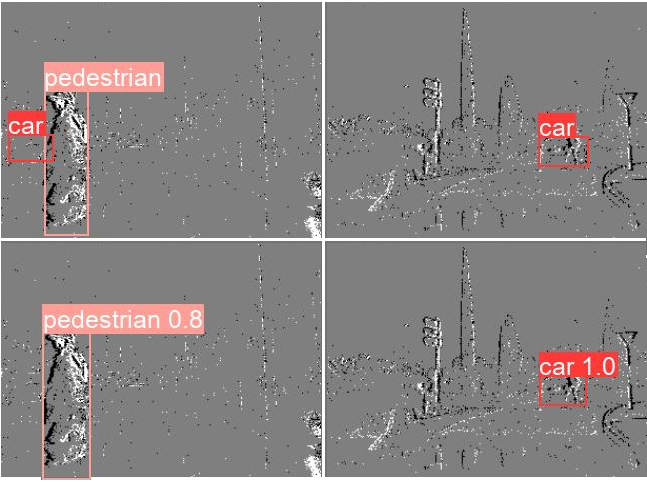}}
  \caption{Example results of object detection using PowerYOLO mixed precision network (bottom) compared to labels (top)}.
  %\caption{Przykładowe wyniki detekcji obiektów za pomocą sieci mieszanej precyzji PowerYOLO (dół) w porównaniu do prawdziwych etykiet (góra)}
  \label{fig:pot2_demo}
\end{figure}

Neural network training was performed based on YOLO8s code available in \cite{b32}, suitably modified to perform quantised training. 
% (all scripts used in the experiments described in this paper, and the quantised models, are available from the external repository \url{github.com/dprze/PowerYOLO}).
The full precision network was trained for 100 epochs with 4 A100 NVIDIA GPUs, using the SGD method, with a momentum of $0.9$ and an initial learning rate of $0.01$ reduced linearly to $0.0001$.
The weights of the full precision network were then used to initialise the quantised model.
The quantised network was trained for 20 epochs, using single A100 NVIDIA GPU, also using SGD and with a low initial learning rate value of $0.0001$.
The value of the learning rate was reduced at epochs 5, 8 and 15 according to $lr=\gamma lr$, where $\gamma=0.1$.
The Exponentially Moving Average (EMA) model was used during training -- the final model parameters were a weighted average of the parameters from each training iteration (this average is quantised logarithmically, and the EMA model is quantised after each update).
In the conducted experiments, it was found that disabling EMA in the quantised learning process results in a decrease in the performance of the network in terms of mAP by several percents.
%Uczenie sieci neuronowych zostało przeprowadzone w oparciu o kod dostępny \cite{b32}, odpowiednio zmodyfikowany do przeprowadzenia uczenia kwantyzowanego (wszelkie skrypty użyte w eksperymentach opisanych w niniejszej pracy, oraz modele kwantyzowane, są dostępne w zewnętrznym repozytorium \url{github.com/dprze/PowerYOLO}).
%Sieć pełnej precyzji uczona jest przez 100 epok metodą SGD, z momentum $0.9$ i początkową wartością współczynnika uczenia $0.01$ redukowaną liniowo do $0.0001$.
%Następnie wagi sieci pełnej precyzji wykorzystywane są do inicjalizacji modelu kwantyzowanego.
%Sieć kwantyzowana uczona jest przez 20 epok, również z użyciem SGD i z niską początkową wartością współczynnika uczenia $0.0001$.
%Wartość współczynnika uczenia redukowana jest w epoce 5, 8 oraz 15 zgodnie z $lr=\gamma lr$, gdzie $\gamma=0.1$.
%W uczeniu wykorzystywany jest model EMA (Exponentially Moving Average) -- ostateczne parametry modelu są średnią ważoną parametrów z każdej iteracji uczenia (średnia ta jest kwantyzowana logarytmicznie, a model EMA jest kwantyzowany po każdej aktualizacji).
%W ramach przeprowadzonych eksperymentów stwierdzono, że wyłączenie EMA w procesie uczenia kwantyzowanego powoduje spadek skuteczności sieci w sensie mAP o kilka procent.
Table \ref{tab:map_results} shows the training results of the Yolov8s network on the Gen1 dataset.
The mAP50 and mAP50-95 are averaged over the two object classes specified: vehicles and pedestrians.
In general, the metric is defined as area under the precision-recall curve and is widely used for evaluation of detection algorithms.
Quantisation to 8-bit integer values after custom QAT PoT training was performed using the OpenVino library (Fig. \ref{fig:pot2_demo} demonstrates the detector operation).

%W tabeli \ref{tab:map_results} pokazano wyniki uczenia sieci Yolov8s na zbiorze Gen1.
%Wskaźniki mAP50 oraz mAP50-95 są uśrednione dla obu wyszczególnianych klas obiektów - pojazdów i pieszych.
%Kwantyzacja do wartości całkowitych ośmiobitowych przeprowadzona została z użyciem biblioteki OpenVino (na Rys. \ref{fig:pot2_demo} zademonstrowano działanie detektora).

\begin{table}[t!]
\small
\caption{Object detection performance for networks: full-precision (baseline), with uniform (INT*) and PoT quantisation (LOG*). 
%\caption{Skuteczność detekcji obiektów dla sieci kolejno pełnej precyzji (baseline), kwantyzowanej do liczb całkowitych ośmiobitowych (INT8 W/A), z wagami w postaci potęg dwójki (LOG4W) oraz z wagami w postaci potęg dwójki i resztą parametrów i aktywacji kwantyzowanymi do liczb całkowitych ośmiobitowych (LOG4W/INT8A).
\label{tab:map_results}}
% \begin{adjustwidth}{-\extralength}{0cm}
\newcolumntype{C}{>{\centering\arraybackslash}X}
\begin{tabularx}{0.49\textwidth}{CCCC}
\toprule
\textbf{Architecture} & \textbf{Quantisation}	& \textbf{mAP50} & \textbf{mAP50-95}\\
\midrule
\multirow[m]{5}{*}{YOLO8s}    & Baseline & 0.586 & 0.340\\
                                & INT8 W/A & 0.582 & 0.330 \\
                                & INT4W & 0.523 & 0.288 \\
                                & LOG4W & 0.566 & 0.312\\
                                & LOG4W/INT8A & 0.554 & 0.301 \\
\bottomrule
\end{tabularx}
\end{table}

A decrease in detection performance is evident for both networks quantised to INT8 and LOG4, with the former being approximately $1\%$ (mAP50), $3\%$ (mAP50-95) and the latter $3.5\%$ (mAP50), $8.3\%$ (mAP50-95) and $5.5\%$ (mAP50), $11.5\%$ (mAP50-95) depending on whether other parameters and activations are also quantised.
Using the same QAT approach, we also trained a model with 4-bit width uniformly quantised weights.
The gap to baseline model is almost 2 times higher than in case of LOG4 model, which shows the superiority of logarithmic quantisation.
However, it's worth noting that there are efforts in the literature to minimise the gap between full precision and low-bitwidth uniformly quantised models by introducing additional mechanisms, like in \cite{b34}.

The comparison with SoTA has to be done in two steps, as to the authors' knowledge there are no other solutions that process event data using YOLO networks with very low computational precision.
Thus, it is first necessary to refer to other object detectors developed on the Gen1 dataset.
A comparison with the full-precision SoTA for the Gen1 dataset is shown in Table \ref{tab:fp32_sota}.
To enable the best possible comparison, we have trained the YOLOv8l network in addition to the YOLOv8s network.
The solution proposed in \cite{b30} uses the YOLOv5l network with a slightly different representation of the event data.
Without the additional mechanisms proposed by the authors, and therefore with the difference only due to the way the data are processed before and after the neural network, and the way the event data are represented, the difference is small, less than $2\%$ for the mAP50-95 index in favour of \cite{b30}.
On the other hand, considering a model with Hyper Histogram event representation and with data augmentation adapted to the event representation, the difference is more than $20\%$.

\begin{table*}[t!]
\caption{Comparison with SoTA full precision for Gen1 dataset against mAP and model size. %highlighting the main differences in architecture and learning data processing.
\label{tab:fp32_sota}}
%\caption{Porównanie z SOTA pełnej precyzji dla zbioru Gen1 względem mAP oraz rozmiaru modelu, z wyróżnieniem głównych różnic w architekturze i przetworzeniu danych uczących. \label{tab:fp32_sota}}
\newcolumntype{C}{>{\centering\arraybackslash}X}
\begin{tabularx}{\textwidth}{CCCCC}
\toprule
Solution & Event representation & Architecture & mAP50-95 & No Parameters\\
\midrule
\multirow[m]{2}{*}{Ours} & \multirow[m]{2}{*}{Event frame} & YOLOv8s & 0.347 & 11.2M \\
& & YOLOv8l & 0.375 & 43.7M \\
 Baseline \cite{b30} & ITS & \multirow[m]{2}{*}{YOLOv5l} & 0.382 & \multirow[m]{2}{*}{46.5M} \\
 AEC \cite{b30} & HH + SM & & 0.470 &  \\
\bottomrule
\end{tabularx}
\end{table*}

Secondly, low-precision YOLO networks should also be analysed. Table \ref{tab:quant_sota} compares our solution with other 4-bit detectors in which all convolution layer weights except the first and last are quantised. %, and in the case of \cite{b34} additionally activations are also quantised.
It is worth noting that in \cite{b22} the solution is based on Post Training Quantisation, which, as a rule, with such low precision computatnois, does not allow to maintain satisfactory performance.
In \cite{b34}, the authors also quantised the activations, so a direct comparison to our model in any version is not possible -- even comparing to the mixed-precision version, for which in our case the difference to the floating-point version is about $11.5\%$, it is hard to say unequivocally which model performs better.
In the case of the mixed-precision version, we additionally quantise the bias and the batch normalisation parameters, which has, as the results presented in Table \ref{tab:map_results} indicate, a major impact.

%Po drugie należy też przeanalizować sieci YOLO niskiej precyzji. W tabeli \ref{tab:quant_sota} zestawiono nasze rozwiązanie z innymi 4-bitowymi detektorami, w których kwantyzacji podlegają wszystkie wagi warstw konwolucyjnych oprócz pierwszej i ostatniej, a w przypadku \cite{b34} dodatkowo aktywacje.
%Warto zaznaczyć, że w \cite{b22} rozwiązanie oparte jest o Post Training Quantization, co z reguły przy tak niskiej precyzji obliczeń nie pozwala na utrzymanie satysfakcjonującej skuteczności działania.
%W \cite{b34} autorzy poddali kwantyzacji również aktywacje, dlatego bezpośrednie porównanie do naszego modelu w żadnej wersji nie jest możliwe - nawet porównując do wersji mixed precision, dla której w naszym przypadku różnica względem wersji zmiennoprzecinkowej wynosi ok. $11.5\%$, ciężko jednoznacznie stwierdzić, który model wypada lepiej.
%W przypadku wersji mixed-precision dokonujemy dodatkowo kwantyzacji biasu oraz parametrów normalizacji partiami, co ma (jak pokazują wyniki zaprezentowane w tabeli \ref{tab:map_results}) duże znaczenie.

Nevertheless, our solution is comprehensive, i.e. it reduces computation on floating-point numbers to a minimum (actually limiting it only to re-quantisation operations, which is unavoidable with any quantisation) and targets all elements (all layers) of the information flow in the neural network.
For this reason, we believe that, in terms of ultimate efficiency with respect to memory-computational complexity, our solution is at least as good as \cite{b34}, especially given the potential for reducing computational complexity introduced by using quantisation for weights with powers of two (as we show in section \ref{subsec:hw_design}), and thus we set a new SoTA in this category.

\begin{table*}[!t]
\caption{Comparison of different SoTAs of low-precision YOLO detectors. It is worth noting that the other methods listed in the table are based on processing classical images and only ours treats event data. 
%All compared methods do not quantise the first and last convolution layers. 
\label{tab:quant_sota}}
%\caption{Porównanie różnych SOTA detektorów Yolo niskiej precyzji. Warto zaznaczyć, że wymienione w tabeli inne metody bazują na przetwarzaniu klasycznych obrazów i jedynie nasza traktuje dane zdarzeniowe. Wszystkie porównywane metody nie kwantyzują pierwszej i ostatniej warstwy konwolucyjnej. \label{tab:quant_sota}}
\newcolumntype{C}{>{\centering\arraybackslash}X}
\begin{tabularx}{\textwidth}{CCCCCC}
\toprule
Solution & Architecture & Dataset & Method & Bitwidth & mAP50-95 gap\\
\midrule
Ours & Yolov8s & Gen1 & PoT QAT, layerwise & 4bits W & 8.3\% \\
{EMA+QC~\cite{b34}} & Yolov5s & COCO & Uni QAT, layerwise & 4bits W\&A & 10.4\% \\
Q-Yolo \cite{b22} & Yolov5s & COCO & Uni PTQ, channelwise & 4bits W & 52.4\% \\
\bottomrule
\end{tabularx}
\end{table*}

Moreover, as was already established in \cite{b3} and \cite{b19}, the hardware implementation of neurons and layers based on BAC (Bitshift Accumulate) units leads to significant area and power reductions.
For a single processing element with weights of bit width 4 and 8-bit activations, using PoT weights allows to increase the power efficiency by a factor of 2.
Using this PE for a complete convolution layer leads to $1.6$ increase of power efficiency and allows for higher frequency operation, as shown for ZCU 104 platform.

\section{Conclusion}

In this work we presented an extremely efficient, mixed precision PowerYOLO detector achieving mAP of $0.301$ for the GEN1 event dataset.
By using both logarithmic quantisation of weights and linear quantisation of activations and other parameters, we eliminated most of floating point operations in the inference flow, allowed replacing multiplication with efficient bit-shifting, and reduced the model size almost 8x while still proposing an accurate detector.
In order to close the gap between floating point solutions, for future work we consider extending the training flow with shadow mosaic augmentation (proper for event data) and using other event representation techniques, possibly allowing to deal with slowly moving and static objects.
% It is also advisable to consider other models.
% Ultimately, we also understand the need to demonstrate the full solution in hardware, showing how the introduced reductions enable the implementation in low-power applications.
% It is important to note that both the low memory and computational complexity of the model, as well as the event source of the input data contributes to energy efficient solution.
Our initial thesis stated that for embedded applications one should address the algorithm, hardware and sensor (data) at once, and with our solution we show that comprehensive analysis of all those elements leads to truly tiny, yet powerful, machine learning solution.

% koniecznosc shadow mosaic, lepszej reprezentacji

% \section*{References}

\end{document}